\documentclass{article}

\usepackage{PRIMEarxiv}
\usepackage{bm}
\usepackage[utf8]{inputenc} 
\usepackage[T1]{fontenc}    
\usepackage{hyperref}       
\usepackage{url}            
\usepackage{booktabs}       
\usepackage{amsfonts}       
\usepackage{amsmath}
\usepackage{amsthm}
\usepackage{nicefrac}       
\usepackage{microtype}      
\usepackage{lipsum}
\usepackage{multirow}
\usepackage{fancyhdr}       
\usepackage{cleveref}
\usepackage{graphicx}       
\newtheorem{theorem}{Theorem}
\newtheorem{proposition}[theorem]{Proposition}
\graphicspath{{media/}}     
\usepackage{subcaption} 
\title{Block Flow: Learning Straight Flow on Data Blocks
}

\author{
  Zibin Wang\\
  East China Normal University\\
   \And
  Zhiyuan Ouyang\\
  East China Normal University\\
  \texttt{a397692218@gmail.com} \\
  \AND
  Xiangyun Zhang \\
  East China Normal University\\
  \texttt{xyzhang@math.ecnu.edu.cn} \\
}

\begin{document}
\maketitle
\begin{abstract}
Flow-matching models provide a powerful framework for various applications, offering efficient sampling and flexible probability path modeling. These models are characterized by flows with low curvature in learned generative trajectories, which results in reduced truncation error at each sampling step. To further reduce curvature, we propose block matching. This novel approach leverages label information to partition the data distribution into blocks and match them with a prior distribution parameterized using the same label information, thereby learning straighter flows. We demonstrate that the variance of the prior distribution can control the curvature upper bound of forward trajectories in flow-matching models. By designing flexible regularization strategies to adjust this variance, we achieve optimal generation performance, effectively balancing the trade-off between maintaining diversity in generated samples and minimizing numerical solver errors. Our results demonstrate competitive performance with models of the same parameter scale.Code is available at \url{https://github.com/wpp13749/block_flow}.
\end{abstract}
\section{Introduction}
Diffusion generative models have emerged as a compelling family of paradigms capable of modeling data distributions by stochastic differential equations (SDEs) \cite{sohl2015deep,ho2020denoising,song2020score},and they have remarkable success in many fields like images generation \cite{rombach2022high,esser2024scaling}, video systhesis\cite{ho2022video,bar2024lumiere}, audio systhesis\cite{kong2020diffwave}, protein design\cite{abramson2024accurate}, and so on. The generative process is defined as the temporal inversion of a forward diffusion process, wherein data is progressively transformed into noise. This approach enables training on a stationary loss function \cite{vincent2011connection}. Moreover, they are not restricted by the invertibility constraint and can generate high-fidelity samples with great diversity, allowing them to be success fully applied to various datasets of unprecedented scales \cite{ramesh2022hierarchical,saharia2022photorealistic}.

Continuous Normalizing Flow (CNF) is defined by \cite{chen2018neural}, which offer the capability to model arbitrary trajectories, encompassing those represented by diffusion processes\cite{song2021maximum}. This approach is particularly appealing as it addresses the suboptimal alignment between noise and data in diffusion models by attempting to build a straight trajectory formulation that directly connects them. With neural ordinary differential equations (ODEs), \cite{lipman2023flowmatchinggenerativemodeling} propose flow-matching to train CNFs and achieve empirically observed improvements in both training efficiency and inference speed compared to traditional diffusion models. 

A key drawback of diffusion/flow-matching models is their high computational cost during inference, as generating a single sample (e.g., an image) requires solving an ODE or SDE using a numerical solver that repeatedly evaluates the computationally expensive neural drift function. \cite{liu2022flowstraightfastlearning} proposes the Rectified Flow to solve this problem with optimal transport framework. In practical applications, the trajectory always deviates from the ideal straight line\cite{nguyen2023bellman}. Therefore, it is necessary to implement additional control measures to ensure the straightness of the trajectory during the operational process.

The straightness of trajectories in Neural ODEs is crucial as it minimizes truncation error in numerical solvers, 
thereby reducing computational costs during sampling. 
Many studies \cite{lee2023minimizing,liu2022flowstraightfastlearning} have made an effort to minimize the curvature of
the learned generative trajectories to achieve a flow model
with a more linear and straighter path. 
We summarize our contributions as follows:
\begin{itemize}
    \item We propose block matching to reduce intersections in forward trajectories by learning 
    straight flows on paired prior distribution blocks and data blocks.
    \item We establish the relationship between the curvature of forward trajectories and the variance of the prior distribution. 
    Based on this proposition, we design a method to reduce the curvature of 
    forward trajectories, enabling the learning of straighter flows without retraining.
    \item We demonstrate the results of our method with multiple strategies, 
    achieving competitive performance with networks of the same parameter scale.
\end{itemize}

\section{Background}
\label{sec: background}
 Let $\mathbb{R}^{d}$ represent the space containing the data points. An interpolation $x_t \sim p(x_t)$ on span $t \in [0,1]$ between empirical observations $x_0 \sim p(x_0)$ and $x_1 \sim p(x_1)$, where $ x_0 $ and $ x_1 $ are of the same dimensionality.
 The diffusion/flow-matching model aim to learn $T_\theta(\cdot)$, which serves as an estimator to map the domain $[0, 1] \times \mathbb{R}^{d}\xrightarrow{}\mathbb{R}^{d}$ that can transport $x_0$ to $x_1$:\\
 \textbf{Diffusion Model\ \ } The diffusion models' goal is to construct a diffusion process to learn the score $s_\theta(x_t,t)$\cite{song2020score}, which can be modeled as the solution to a reverse time SDE:
 \begin{equation}
     dx=[f(x_t,t)-g^2(t)\nabla_{x_t}\log p(x_t)]dt+g(t)d\bar{w},
     \label{SDE}
 \end{equation}
 where $\bar{w}$ is the standard Wiener process, $f(\cdot,t):\mathbb{R}^d \xrightarrow{} \mathbb{R}^d$ is the drift coefficient of $x_t$, and $g(\cdot):\mathbb{R}\xrightarrow{}\mathbb{R}$ is a scalar diffusion coefficient of $x_t$. The training objective is:
\begin{equation}
    \label{diffusion}
    \theta^*=\mathop{\arg\min}\limits_{\theta}
    \int_0^1 \lambda(t) \mathbb{E}_{x_0\sim p(x_0)}\mathbb{E}_{x_t\sim p(x_t|x_0)} [ ||s_{\theta}(x_t, t) - \nabla_{x_t}\log p(x_t|x_0)||_2^2 ] dt.
\end{equation}
where $\lambda(\cdot): [0,1]\xrightarrow{}\mathbb{R}_{>0}$ is a positive function, typically choose $\lambda(t)\propto \frac{1}{\mathbb{E}[||\nabla_{x_t}\log p(x_t|x_0)||_2^2]}$. Given sufficient data and model capacity, the optimal diffusion model $s_{\theta^*}(x_t, t)$ matches $\nabla_{x_t}\log p(x_t|x_0)$. 
 And samples can be generated by starting from $x_1\sim \mathcal{N}(0,I)$.
 In the discrete case, Eq.~\eqref{SDE} is often reformulated by \cite{ho2020denoising} in the following form:
 \begin{equation}
     x_t=\alpha(t)\ x_0+\sqrt{1-\alpha(t)^2}\ x_1,
 \end{equation}
 where $\alpha(t)$ is a predefined nonlinear function, such that $x_t$ is the nonlinear interpolation.\\
\textbf{flow-matching model\ \ } The flow-matching models aim to learn avector field $v_\theta(x_t, t)$, which is an estimator solution of the following ODE:
\begin{equation}
\label{Dynamic System}
dx=v_\theta(x_t, t)dt,
\end{equation}
where $x(\cdot)$, referred to as a flow, represents the solution to Eq. ~\eqref{Dynamic System}, illustrating the trajectory of the ODE originating from the initial point $x_0$. flow-matching's  training objective is the same as solving a simple least squares regression problem:
\begin{equation}
\label{Flowbased}
    \theta^*=\mathop{\arg\min}\limits_{\theta}
    \int_0^1 \mathbb{E}_{x_t\sim p(x_t)} \left[ ||v_{\theta}(x_t, t)- v_t(x_t)||_2^2 \right] dt,\\
\end{equation}
\begin{equation}
    x_t=tx_1+(1-t)x_0, \label{Rectified flows}
\end{equation}
where $x_t$ is the linear interpolation between $x_0$ and $x_1$. For simplicity, $x_t$ follows the ODE described by the optimal transport flow-matching objective $dx_t = (x_1 - x_0)dt$, which exhibits non-causal behavior because the update of $x_t$ relies on information from the end point $x_1$\cite{liu2022flowstraightfastlearning}.

\begin{figure*}[ht]
\vskip 0.1in
\begin{center}
\centerline{\includegraphics[width=0.9\textwidth]{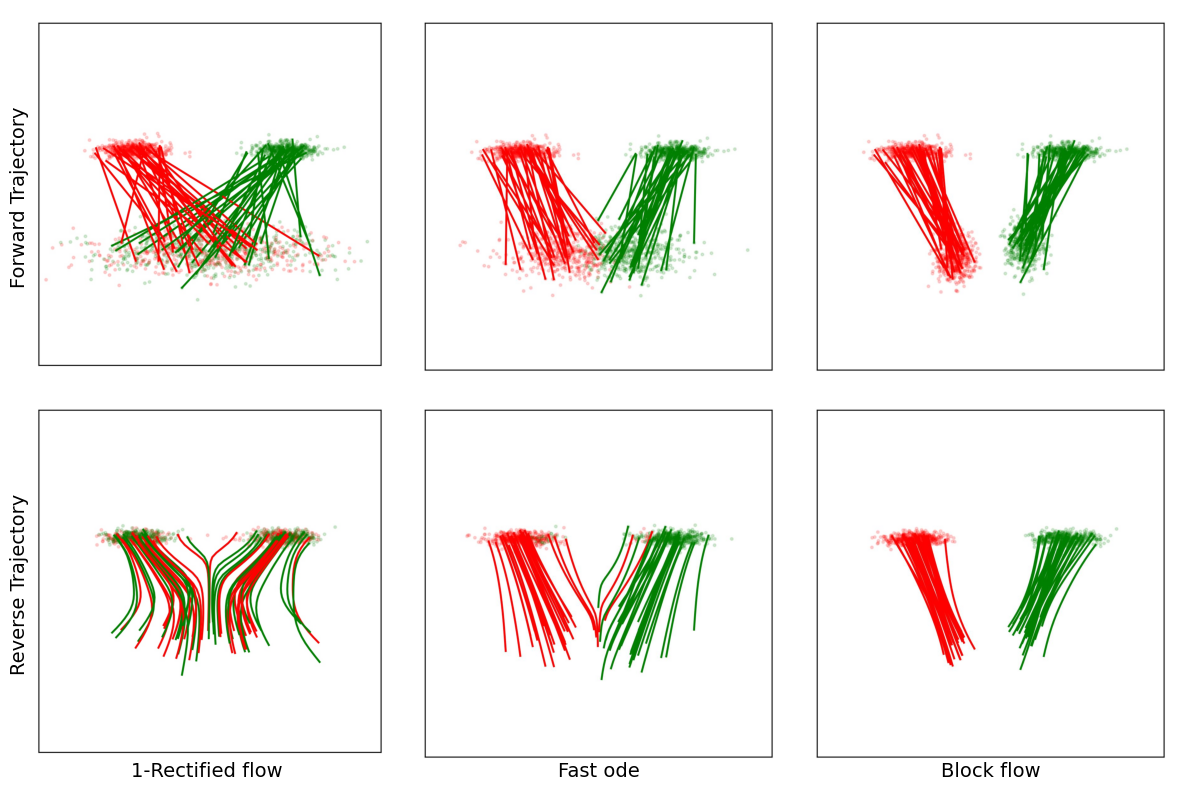}}
\caption{Forward and reverse trajectories of 1-rectified flow 
\cite{liu2022flowstraightfastlearning}, fast ODE \cite{lee2023minimizing}, 
and block flow on a toy example. Intersections in forward trajectories 
lead to averaging effects, causing the resulting reverse trajectories 
to deviate from the true data distribution. 
We partition the toy example into two blocks and match each block with a separate Gaussian distribution.By employing block matching, block flow significantly reduces 
cross-label intersections in forward trajectories, resulting 
in straighter reverse trajectories.}
\label{block_six_fig}
\end{center}
\vskip -0.2in
\end{figure*}

\section{Method}
\label{Method}
\subsection{Block Matching}
\label{Block Matching}
To construct straight flow paths with Eq.~\eqref{Rectified flows},
samples $(x_0, x_1)$ are drawn from the joint distribution $p(x_0, x_1)$.   
When $x_0$ and $x_1$ are independent,
regions with significant differences in the data distribution $p(x_0)$ 
are more likely to be matched to regions with minimal differences in 
the prior distribution $p(x_1)$, leading to increased trajectory intersections. 
These intersections elevate the curvature of the learned generative trajectories. 

We propose leveraging prior knowledge to partition the data distribution 
into blocks, with each block matching localized regions of the prior 
distribution. We term this matching strategy block matching and parameterize 
the prior distribution to make block matching learnable. 
As showing in Fig.~\ref{block_six_fig}, block matching ensures that proximate 
regions of the data distribution $p(x_0)$ are matched with proximate regions 
of the prior distribution $p(x_1)$, thereby reducing intersections in forward 
trajectories. 

The implementation of block matching can be divided into three aspects:
\begin{itemize}
    \item Partitioning the data distribution into blocks using readily available label information. 
    For each label $y$, the data block is defined as the conditional distribution $ {p(x_0\mid y)}$.  
    \item Ensuring that the localized region of the prior distribution matched to each data block $p(x_0\mid y)$ is simple and controllable. 
    Specifically, the localized region of the prior distribution, constructed with label $ y $, 
    follows a Gaussian distribution $ \mathcal{N}(\mu_y, \Sigma_y) $. In this case, the prior distribution is a mixture of Gaussians  
    defined as $ p(z) =\int p(y) p(z\mid y)dy $, where $z \mid y\sim \mathcal{N}(\mu_y, \Sigma_y)$, $ p(y) $ can be interpreted as the mixture coefficient to $ p(z \mid y) $. 
    \item $ (\mu_y, \Sigma_y) $ are learnable parameters. Following \cite{lee2023minimizing},
    we employ a small-size encoder to obtain $\left(\mu_{\phi(y)}, \Sigma_{\phi(y)}\right)$.Given a label $ y $, 
    flow-matching is performed between the Gaussian distribution $ q_{\phi}(z \mid y) $ and the data block $x_0\mid y$. 
\end{itemize}

\subsection{Curvature Control for Forward Trajectories}
\label{Curvature Control for Forward Trajectories}
Reducing the curvature of flow paths in Neural ODE-based models can decrease the error of numerical solvers. 
When the curvature is small, the flow paths become straighter, enabling accurate results with fewer steps during sampling. 
The curvature of the generative process $ x_t $, derived from $(x_0, x_1)$ 
based on Eq.~\eqref{Rectified flows} , is defined as:  
\begin{equation}
V((x_0, x_1)) = \int_0^1 \mathbb{E}_{(x_0,x_1)\sim p(x_0, x_1)} \bigg[ \left\| (x_1 - x_0) \right. \left. - \mathbb{E}[x_1 - x_0 \mid x_t] \right\|^2_2 \bigg] dt,
\label{Rectified flows Curvature}
\end{equation}
which is equal to the straightness measure introduced in \cite{liu2022flowstraightfastlearning}. 
Here, $ x_0 $ and $ x_1 $ can be either independent or coupled. 
To identify an appropriate prior distribution $ p(x_1) $ that minimizes $ V((x_0, x_1)) $ while preserving the original data distribution, 
we introduce the following proposition:
\begin{proposition}
\label{prop:v=0} 
If the probability distribution of $ x_c $ is a Dirac delta function, i.e.,  
$x_c \sim \delta(x - c),$ then:  
\begin{equation}
    V((x_0, x_c))=0. 
\end{equation}
\end{proposition}
\cref{prop:v=0} indicates that finding a simple prior distribution (e.g., Dirac delta function) 
to achieve $ V((x_0, x_1))=0 $ is straightforward. 
However, generating high-quality images from a Dirac delta function remains inherently challenging.
To achieve precise control over the curvature of the forward trajectories 
while ensure generation quality, we introduce the following proposition:
\begin{proposition}
\label{prop:v<var} 
For any joint distribution $p(x_0, x_1)$, the following holds:  
\begin{equation}
V((x_0, x_1))\leq \left( \sqrt{\text{Var}(x_1)} + \sqrt{\text{Var}(x_0)} \right)^2.    
\end{equation}
In particular, when $ x_0 $ and $ x_1 $ are independent, we have:  
\begin{equation}
    V((x_0, x_1))\leq \text{Var}(x_1) + \text{Var}(x_0).
\end{equation}

\end{proposition}
According to \cref{prop:v<var}, the upper bound of $ V((x_0, x_1)) $ can be controlled by adjusting 
the variance of $ p(x_1) $ without modifying the data distribution $ p(x_0) $. As outlined in \cref{Block Matching},when $ x_1 \sim p(z) =\int p(y) p_{\phi}(z \mid y)dy $ is parameterized as a mixture of Gaussians, 
the mean and covariance \cite{murphy2012machine} of $ x_1 $ is given by: 
\begin{equation}
    \mathbb{E}[x_1]=\int p(y)\mu_{\phi(y)}dy,
\end{equation}

\begin{align}
\text{Cov}[x_1] &= \int p(y)\left[\Sigma_{\phi(y)} + \mu_{\phi(y)}\mu_{\phi(y)}^T\right] dy - \mathbb{E}[x_1]\mathbb{E}[x_1]^T \nonumber \\
&= \mathbb{E}_y\left[\Sigma_{\phi(y)} + \mu_{\phi(y)}\mu_{\phi(y)}^T\right] - \mathbb{E}[x_1]\mathbb{E}[x_1]^T.
\end{align}
By constraining the covariance $\Sigma_{\phi(y)}$, the within-group variance $\mathbb{E}_y[\Sigma_{\phi(y)}]$ of the mixture of Gaussians is reduced. 
Furthermore, limiting the range of the means $\mu_{\phi(y)}$ provides additional control over the between-group variance $\mathbb{E}_y[\mu_{\phi(y)}\mu_{\phi(y)}^T]-\mathbb{E}[x_1]\mathbb{E}[x_1]^T$. 
Results in \cref{Experiment Detail} and \cref{tab: priorvar} demonstrate that the constraints on the means $\mu_{\phi(y)}$ can be safely ignored, 
as the between-group variance accounts for only a small fraction of the total variance.

\begin{figure}[ht]
\vskip 0.2in
\begin{center}
\centerline{\includegraphics[width=\columnwidth]{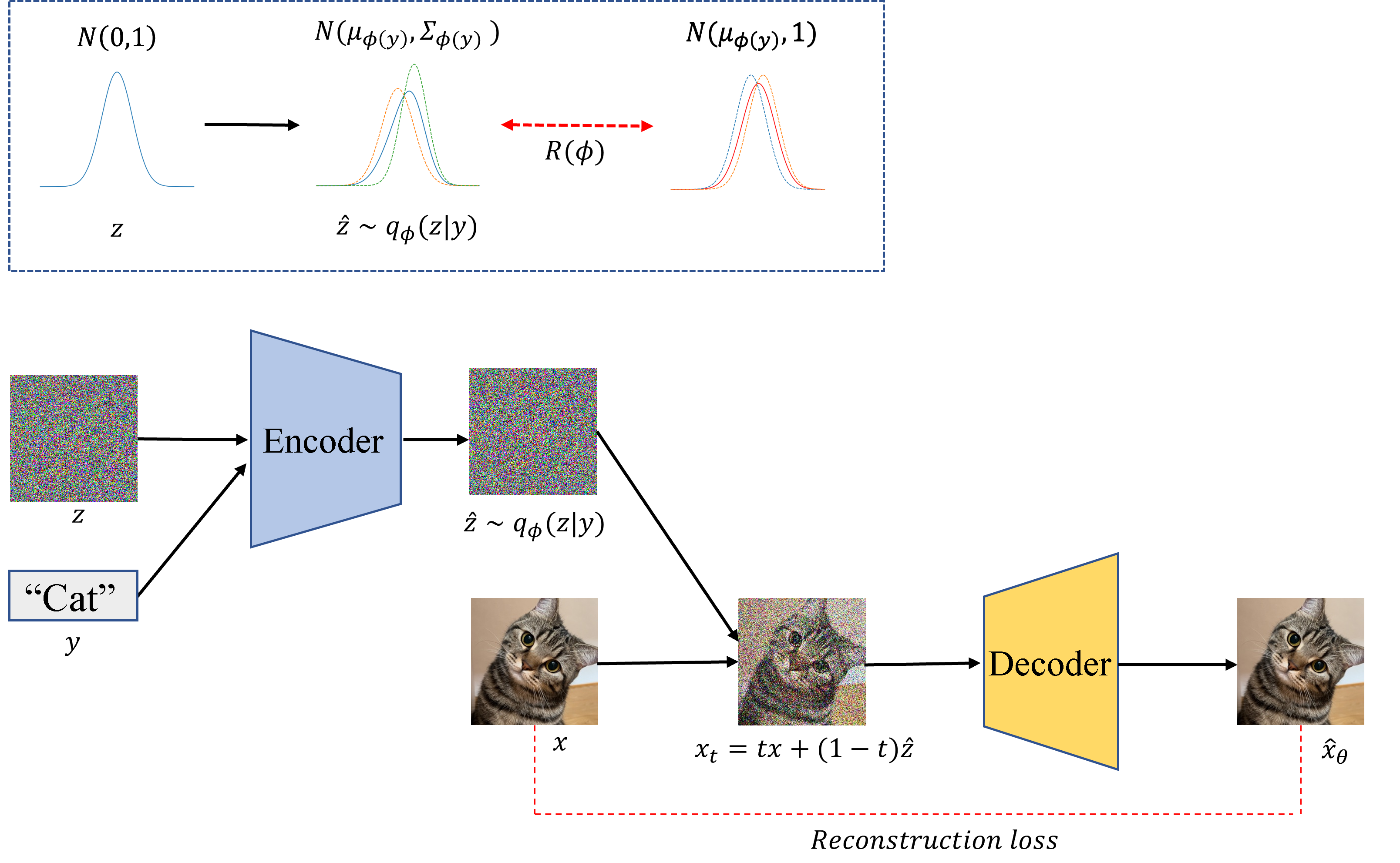}}
\caption{A visualization of the block flow training}
\label{visual_schematic}
\end{center}
\vskip -0.2in
\end{figure}

\subsection{Regularization Strategy in Training}
\label{Regularization Strategy in Training}
The Block flow sampling process begins with a Gaussian mixture distribution $x_1 \sim q_{\phi}(z) =\int p(y) q_{\phi}(z \mid y)dy $, 
and the corresponding training process must align with this sampling procedure.
The sampling process is implemented as \cref{alg:block_flow Sampling}.
Depending on the type of information utilized, two distinct training schemes are proposed. Each scheme is associated with its corresponding regularization strategies, which constrain the learnable parameters $\left(\mu_{\phi(y)}, \Sigma_{\phi(y)}\right)$ 
to prevent the collapse of the prior distribution. Regardless of the regularization strategy, the loss function takes the following form:

\begin{equation}
    \min_{\theta, \phi} \mathbb{E}_{t, (x,y,z) \sim q_{\phi}(x,y,z)} 
    \Bigl[ \left\| (x-z) - v_{\theta}(x_t, t) \right\|^2_2  + \beta R(\phi) \Bigr],
\end{equation}

where $R(\phi)$ denotes the regularization term, and $ q_{\phi}(x,y,z) $ represents the parameterized joint distribution of $ x $, $ y $, and $ z $.The training process is implemented as \cref{alg:Block Flow Training} and Fig.~\ref{visual_schematic}.

\begin{algorithm}[tb]
    \caption{Block Flow Sampling with Euler}
    \label{alg:block_flow Sampling}
\begin{algorithmic}[1]
    \STATE $y_0 \sim p_{data}(y_0), z \sim \mathcal{N}(\bm{0}, \bm{I})$
    \STATE $x_1 = \mu_\phi(y_0) + \Sigma_\phi(y_0) z$
    \FOR{$t = 1, \frac{N-1}{N}, \frac{N-2}{N}, \ldots, 0$}
    \STATE $x_{t-1}=x_t-\frac{1}{N} v_{\theta}(x_t, t)$
    \ENDFOR
    \STATE \textbf{return} $x_0$
\end{algorithmic}
\end{algorithm}
\begin{algorithm}[tb]
  \caption{Block Flow Training}
  \label{alg:Block Flow Training}
\begin{algorithmic}[1]
  \REPEAT
  \STATE $(x_0,y_0)\sim p_{data}\left(x_0,y_0\right)$
  \STATE $t \sim \mathcal{U}(0, 1)$
  \STATE $z\sim \mathcal{N}(\bm{0}, \bm{I})$
  \STATE $z_y=\mu_{\phi (y_0)}+\Sigma_{\phi (y_0)} z$
  \STATE $x_t=tx_0+(1-t)z_y$
  \STATE Take gradient descent step on the model parameters:\\ 
         $\nabla_{\phi,\theta}\left[ \left\| (x_0-z_y) - v_{\theta}(x_t, t) \right\|^2_2 + 
          \beta R(\phi) \right]$
  \UNTIL converged
\end{algorithmic}
\end{algorithm}

\subsubsection{Full Alignment Scheme}
\label{Full Alignment Scheme}
The full alignment scheme encodes the label $ y $ to obtain $ (\mu_{\phi(y)},\Sigma_{\phi(y)}) $, which is then matched with the corresponding data block.Both the encoder's training and sampling processes rely exclusively on the label $ y $ information,
 ensuring complete alignment between these phases. During training, two regularization methods can be applied:  
\begin{itemize}
    \item Norm Regularization: Constrains the norm of $ \log \Sigma_{\phi(y)} $ without restricting $ \mu_y $. The corresponding regularization term is:  
    \begin{equation}
        R(\phi)=\left\| \log \Sigma_{\phi(y)}  \right\|.
    \end{equation}
    \item $\beta$-VAE Regularization:Controls the KL divergence between $ \mathcal{N}\left(\mu_{\phi(y)} , \Sigma_{\phi(y)}\right) $ and $ \mathcal{N}(\bm{0}, \bm{I}) $, imposing constraints on both $ \mu_y$ and $ \Sigma_{\phi(y)} $. The corresponding regularization term is:  
    \begin{equation}
    R(\phi)=D_{KL} \left( q_{\phi}(z \mid y) || p_0(z) \right).
    \end{equation}
    The corresponding loss function resembles the $\beta$-VAE objective.
\end{itemize}

\subsubsection{Hybrid Information Alignment Scheme}
\label{Hybrid Information Alignment Scheme}
To leverage the intrinsic relationship between the dataset samples $ x $ and labels $ y $, 
the hybrid information matching scheme encodes $ x $ and $ y $ to 
obtain  $ (\mu_{\phi(x,y)},\Sigma_{\phi(x,y)}) $, ensuring that the starting point of the flow 
is closer to the data distribution. While this is straightforward during training, 
relying on $ x $ during sampling would require providing existing samples to 
generate new ones, which is a restrictive condition. 
Thus, the sampling process relies exclusively on the label $ y $, starting from the Gaussian mixture distribution $ q_{\phi}(z) =\int p(y) q_{\phi}(z \mid y)dy $. This creates a misalignment between the training and sampling phases. To address this, the following two regularization methods are introduced, which also help reduce the misalignment:
\begin{itemize}
    \item Conditional $\beta$-VAE Regularization: The encoder uses the label $ y $ to obtain $ (\mu_{\phi(y)},\Sigma_{\phi(y)}) $, and the KL divergence between $ \mathcal{N}\left(\mu_{\phi(x,y)},\Sigma_{\phi(x,y)}\right)  $ and $ \mathcal{N}\left(\mu_{\phi(y)} , \Sigma_{\phi(y)}\right) $ is minimized to reduce the discrepancy between the prior distributions derived from hybrid information and label-only information. The corresponding regularization term is:  
    \begin{equation}
        R(\phi)=D_{\text{KL}} \left( q_{\phi}(z \mid x, y) || p_{\theta}(z \mid y) \right).
    \end{equation}
    
    The corresponding loss function resembles the conditional $\beta$-VAE objective.
    \item $\beta$-VAE Regularization:In training process, the encoder uses both the sample $ x $ and the label $ y $ to obtain $ \left(\mu_{\phi(x,y)}, \Sigma_{\phi(x,y)}\right)$. With a certain probability, $ x $ is dropped out, causing $ q_\phi(z | x, y) $ to reduce to $ q_\phi(z | y) $. The KL divergence between $ \mathcal{N}\left(\mu_{\phi(x,y)},\Sigma_{\phi(x,y)}\right)  $ and $ \mathcal{N}(\bm{0}, \bm{I}) $ is controlled. The corresponding regularization term is: \begin{equation}
        R(\phi)=D_{\text{KL}} \left( q_{\phi}(z \mid x, y) || p_0(z) \right).    
    \end{equation}   
\end{itemize}

\begin{table*}[t]
\begin{center}
\begin{tabular}{p{7cm}ccc}
\toprule 
\multicolumn{1}{l}{\textbf{Method}} & \multicolumn{1}{c}{\textbf{NFEs}} & \multicolumn{1}{c}{\textbf{IS ($\uparrow$)}} & \multicolumn{1}{c}{\textbf{FID ($\downarrow$)}} \\
\midrule 
\multicolumn{4}{l}{\textit{GANs}} \\
\midrule
StyleGAN2 \cite{karras2020training} & 1 & 9.82 & 8.32 \\
StyleGAN2 $+$ ADA \cite{karras2020training} & 1 & 9.40 & 2.92 \\
StyleGAN2 $+$ DiffAug \cite{zhao2020differentiable} & 1 & 9.40 & 5.79 \\
TransGAN $+$ DiffAug \cite{jiang2021transgan} & 1 & 9.02 & 9.26 \\
\midrule
\multicolumn{4}{l}{\textit{ODE/SDE-based models}} \\
\midrule
DDPM \cite{ho2020denoising}            &  1000   &  9.41  &  3.21  \\ 
NCSN++ (VE ODE) \cite{song2020score}            &  2000   &  9.83  &  2.38  \\ 
VP ODE             &  140   &  9.37  &  3.93  \\ 
Rectified Flow \cite{liu2022flowstraightfastlearning}  &  127  &  9.60  &  2.58  \\
Fast Ode (RK45)\cite{lee2023minimizing}            &  118   &  9.55  &  2.45  \\ 
Fast Ode (Euler) & 8 & 8.4 & 13.52\\
BOSS \cite{nguyen2023bellman} & 8 & - & 15.74\\
SDM \cite{cao2024spiking} & 8 & 7.84 & 12.95\\

\midrule
\multicolumn{4}{l}{\textit{Ours ($\beta=1$)}} \\ 
\midrule
FANR (RK45) & 117 & 9.61 & 2.29\\
FABR (RK45) & 113 & 9.66 & 2.29\\
HACBR (RK45) & 115 & 9.64 & 2.34\\
HABR (RK45) & 112 & 9.69& 2.30\\
FANR (Euler) & 8 & 8.43 & 13.43 \\
FABR (Euler) & 8 & 8.49 & 12.95 \\
HACBR (Euler) & 8 & 8.43 & 13.52\\
HABR (Euler) & 8 & 8.57 & 12.95\\
\bottomrule   
\end{tabular}
\end{center}
\caption{Comparisons with the state-of-the-art on CIFAR-10 dataset}
\label{results-cifa10}
\end{table*}

\section{Related Works}
\label{Related Works}
\textbf{Different Prior Distributions\ \ } PFGM++ \cite{xu2023pfgm++} extends the flow from the data distribution to infinity, forming a uniform distribution. Sampling starts from this uniform distribution. \cite{jia2024struct} adopt a Gaussian mixture as the prior distribution, treating the associated parameters as hyperparameters. However, they do not provide a strategy for selecting these hyperparameters. In our work, the prior distribution is a Gaussian distribution with learnable parameters, which can be indirectly controlled by adjusting the regularization coefficients.

\textbf{Parameterized Joint Distribution\ \ } \cite{lee2023minimizing} 
parameterizes the joint distribution $q(x, z)$ using a neural network. However, this approach leads to a misalignment between the training and sampling processes: training is based on the parameterized prior distribution $q_{\phi}(z \mid x) $, while sampling starts from
$\mathcal{N}(\bm{0}, \bm{I})$. The misalignment led \cite{lee2023minimizing} to recommend the use of larger regularization coefficients
to force the Gaussian distribution during training to approximate closely $\mathcal{N}(\bm{0}, \bm{I})$. 
In our work, we leverage the label $ y $ to ensure that both training and sampling start from a Gaussian mixture. Furthermore,
in \cref{Regularization Strategy in Training}, we propose multiple schemes to align the training and sampling processes. 
This allows for more flexible choices of regularization coefficients and improves generation quality.

\textbf{Straightness of flow\ \ } \cite{liu2022flowstraightfastlearning} proposed a technique called reflow, which optimizes probability 
sampling paths by retraining the model to achieve straighter flow paths. 
This optimization not only reduces the number of function evaluations (NFE) during sampling but also enhances the quality of the generated images. Later,\cite{liu2022flowstraightfastlearning} extended this framework to larger datasets of high-resolution images, 
achieving impressive results. Additionally, \cite{nguyen2023bellman} employed the Bellman Optimal Stepsize Straightening (BOSS) 
technique to distill flow-matching generative models by segmenting the flow and straightening each segment.

\section{Experiment}
\label{Experiment}
\textbf{Setting\ \ } We evaluate our method on image generation task on the CIFAR-10 and MNIST datasets. 
Following \cite{lee2023minimizing},
    we employ a small-size encoder to obtain $\left(\mu_{\phi(y)}, \Sigma_{\phi(y)}\right)$ and adopt the U-Net architecture from DDPM++ \cite{song2020score} to estimate the vector field $ v_\theta $. 
\cref{results-cifa10} presents the results of four regularization strategies on CIFAR-10 dataset:  
\begin{itemize}
  \item Full Alignment Scheme with Norm Regularization (FANR): 
  In the regularization term, the $L_1$ norm is utilized for $\left\| \log \Sigma_{\phi(y)}  \right\|$
  \item Full Alignment Scheme with $\beta$-VAE Regularization (FABR)
  \item Hybrid Information Alignment Scheme with Conditional$\beta$-VAE Regularization (HACBR)
  \item Hybrid Information Alignment Scheme with $\beta$-VAE Regularization (HABR)
\end{itemize}
When presenting variance results, 
we compute the $L_1$ norm of the corresponding tensor by default. 
To estimate $V((x_0, x_1))$ for each reverse trajectory, 
we use the output of the neural network $ v_{\theta}(x_t, t) $ as a substitute for $\mathbb{E}[x_1 - x_0 \mid x_t]$ in Eq. 
The final $V((x_0, x_1))$ is estimated using 500,000 reverse trajectories generated by a 20-step Euler solver. For more details, refer to \cref{Experiment Detail}.  

Regardless of the regularization strategy, comparative results can be achieved with an appropriate choice of $ \beta $. All four regularization strategies achieve favorable results at $ \beta=1 $, which reflects a trade-off between the variance of the prior distribution (or the curvature of reverse trajectories) and the image generation quality.

\begin{figure}[htbp]
    \centering
    \begin{minipage}[b]{0.45\linewidth}  
        \centering
        \includegraphics[width=0.9\linewidth]{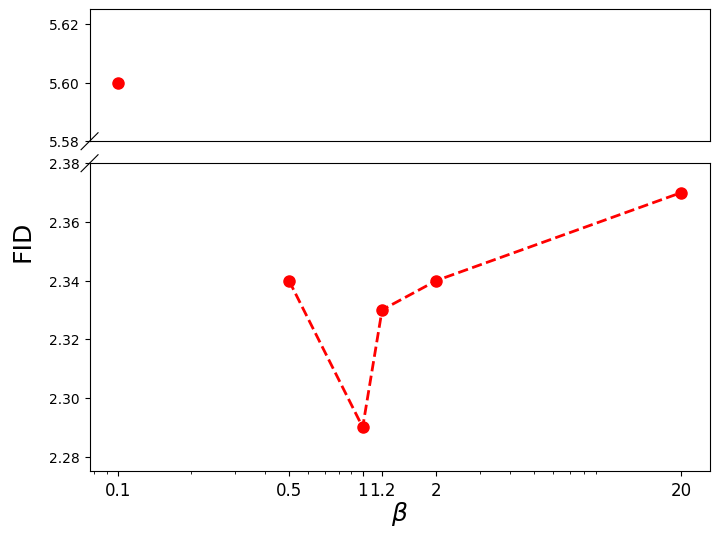}  
        \scriptsize\textbf{(a) $ \beta $ vs FID}  
        \label{subplot_a}  
    \end{minipage}
    \hfill
    \begin{minipage}[b]{0.45\linewidth}
        \centering
        \includegraphics[width=0.9\linewidth]{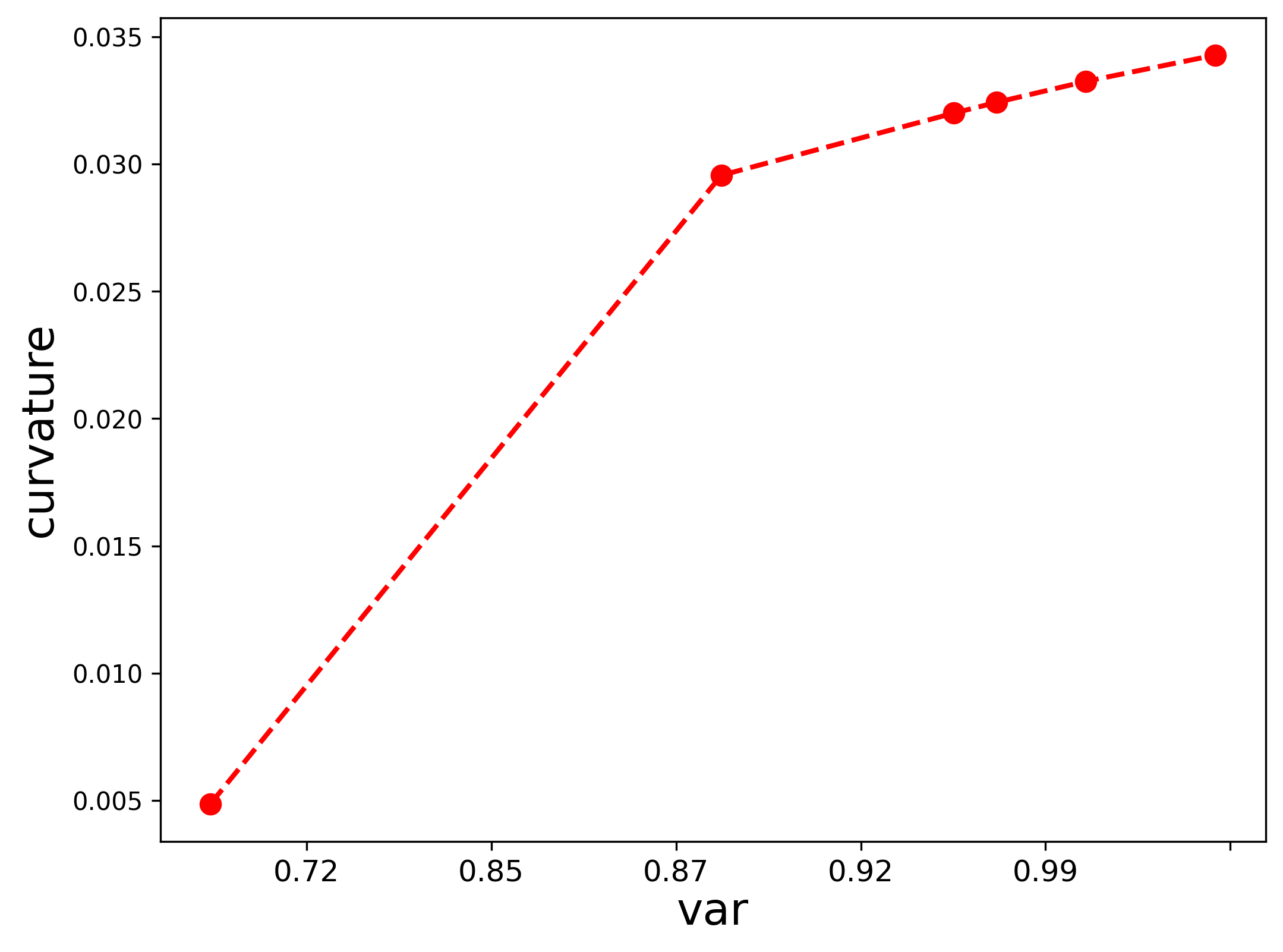}
        \scriptsize\textbf{(b) variance vs curvature}
        \label{subplot_b}
    \end{minipage}
    
    
    \begin{minipage}[b]{0.45\linewidth}
        \centering
        \includegraphics[width=0.9\linewidth]{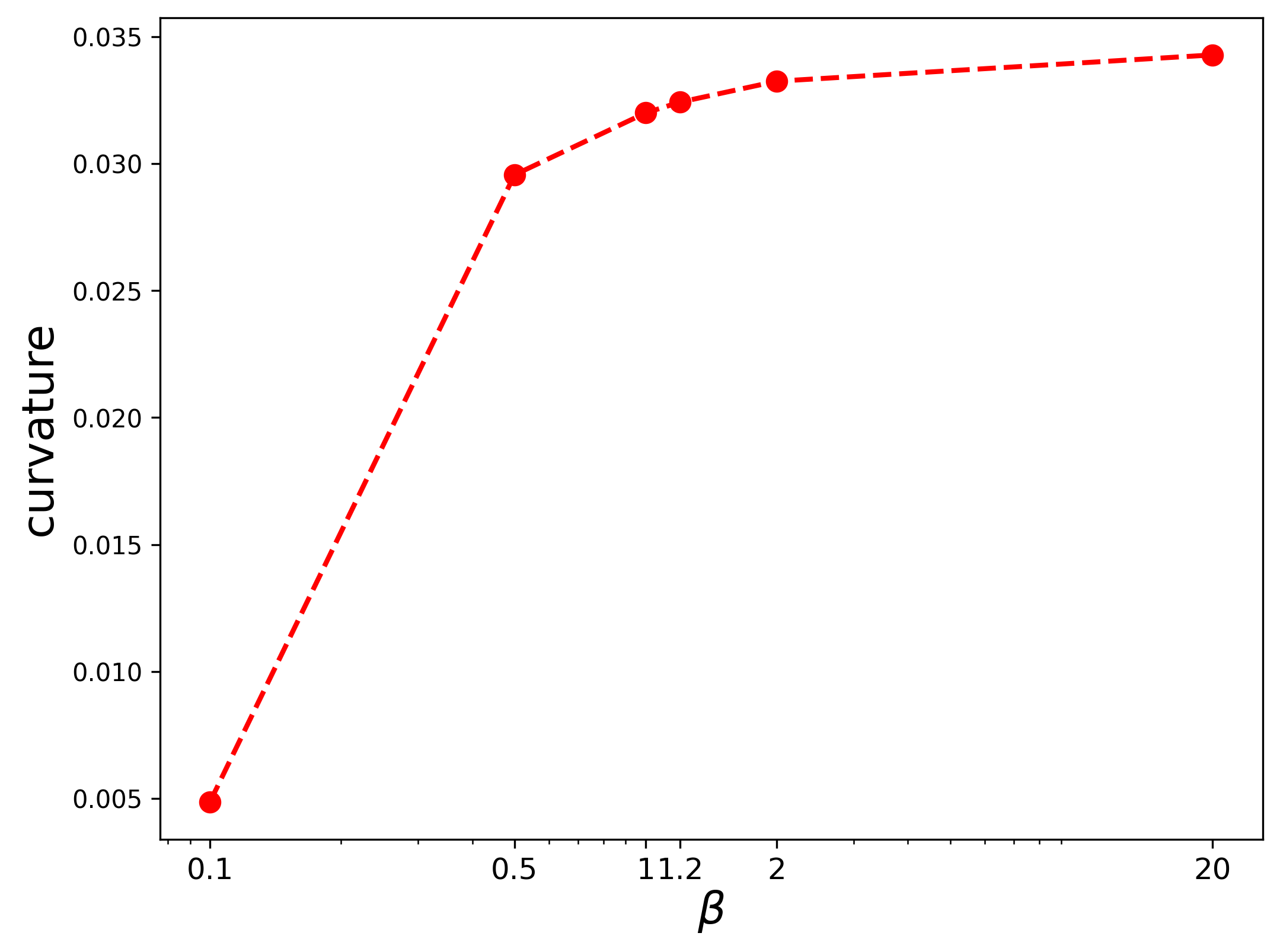}
        \scriptsize\textbf{(c) $ \beta $ vs curvature}
        \label{subplot_c}  
    \end{minipage}
    \hfill
    \begin{minipage}[b]{0.45\linewidth}
        \centering
        \includegraphics[width=0.9\linewidth]{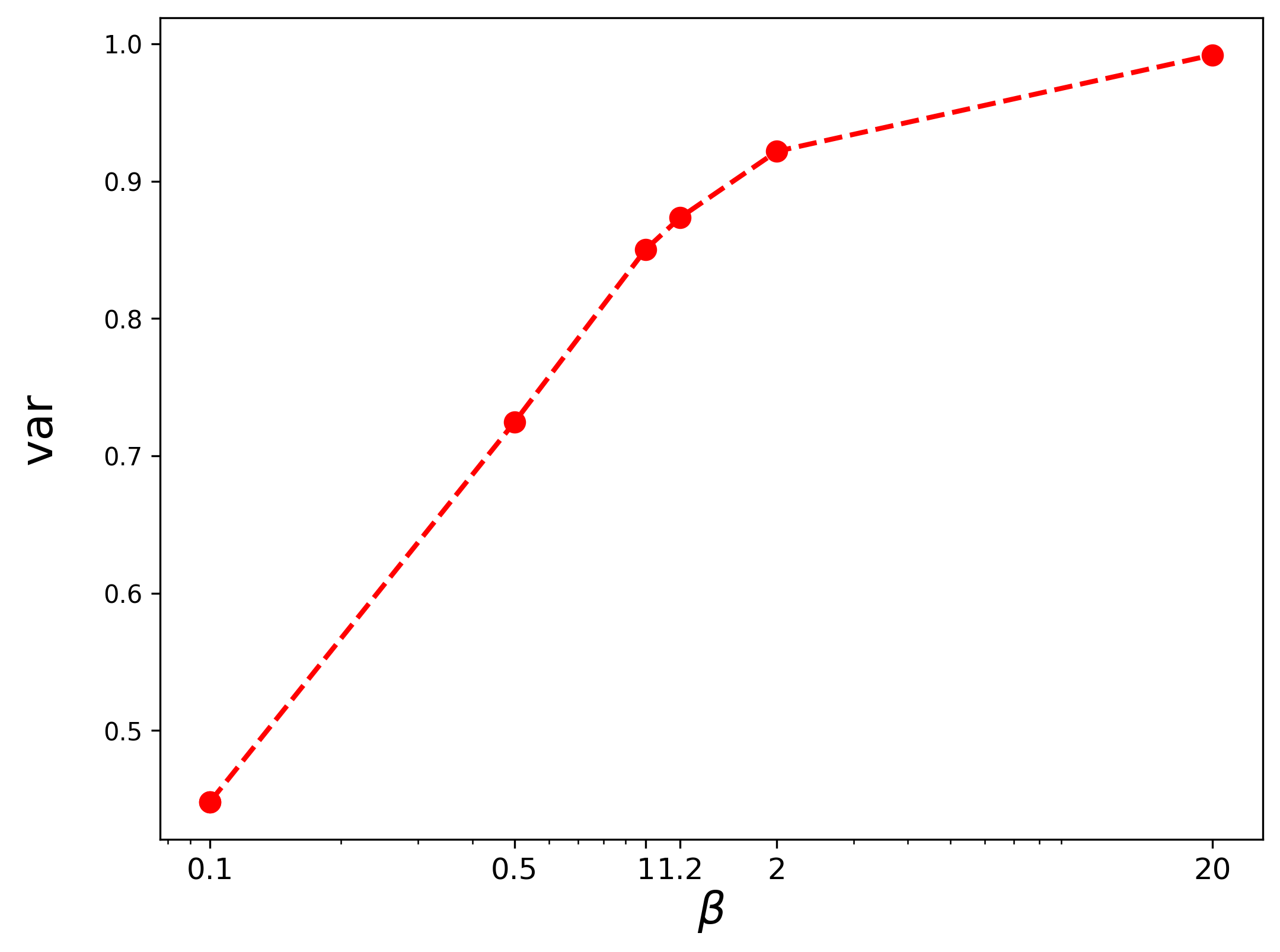}
        \scriptsize\textbf{(d) $ \beta $ vs variance}
        \label{subplot_d}
    \end{minipage}
    \caption{Quantitative results of FABR with varying $beta$ on the CIFAR-10 dataset.}
    \label{4_subplots} 
\end{figure}

\begin{figure}[h]
    \centering
    \begin{subfigure}[b]{0.48\columnwidth} 
        \includegraphics[width=\textwidth]{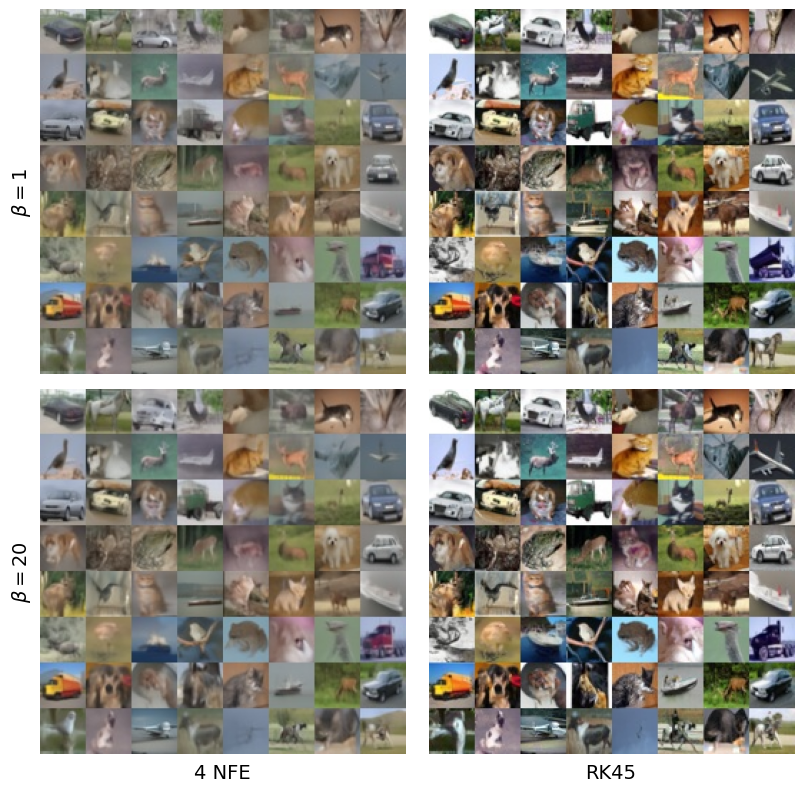}
        \caption{Block Flow (HABR) with Different $\beta$ on the CIFAR-10 dataset using Euler (4 NFEs) and Adaptive-Step RK45 Solvers}
        \label{fig:cifar10}
    \end{subfigure}
    \hfill 
    \begin{subfigure}[b]{0.48\columnwidth} 
        \includegraphics[width=\textwidth]{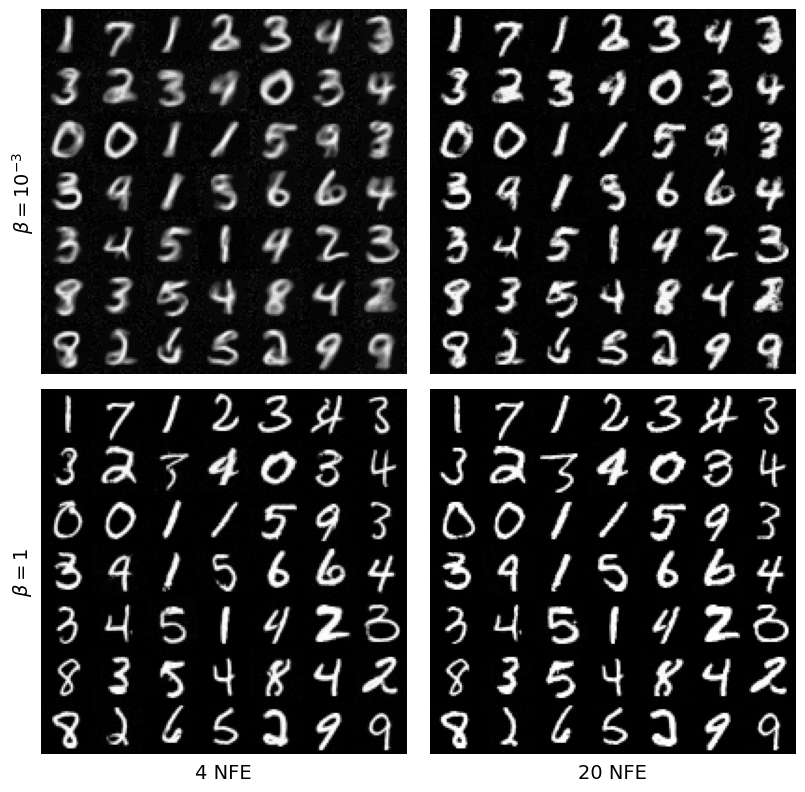}
        \caption{Block flow (FANR) with Different $\beta$ on MNIST dataset using Euler}
        \label{fig:mnist}
    \end{subfigure}
    \caption{Comparison of Block Flow on CIFAR-10 and MNIST datasets} 
    \label{fig:comparison}
\end{figure}

Taking the FABR as an example, as shown in Fig.~\ref{4_subplots}, $ \beta $, 
the variance of the prior distribution, and the curvature of reverse trajectories 
exhibit pairwise positive correlations. 
When $ \beta $ exceeds 1, the growth of curvature and variance slows down significantly. 
$ \beta=1 $ is not only the point where generation quality is optimal but also the inflection point 
for changes in variance and curvature.  
When $ \beta $ is too small (e.g., $ \beta=0.1 $), 
the prior distribution collapses, resulting in a lack of diversity in the generated samples. 
Conversely,when $ \beta $ is too large, the increased curvature introduces higher truncation errors 
in the numerical solver. For a visual comparison of different $ \beta $ in other strategies, see Fig.~\ref{fig:comparison}.  

\begin{table}[H]
    \centering
    \begin{tabular}{lccc}
        \toprule
        Strategy &  $\beta$ & Ratio & Total Var\\
        \midrule
        FABR & 0.1 & 2.5$\times$10$^{-5}$ & 0.4480\\
             & 1 & 1.3$\times$10$^{-7}$ & 0.8503\\
             & 20 & 3.0$\times$10$^{-10}$ &  0.9919\\
        \cmidrule(r){1-4}
        FANR & 0.1 & 2.0$\times$10$^{-4}$ & 0.9796\\
             & 1 & 1.0$\times$10$^{-11}$ & 1.0000\\
        \cmidrule(r){1-4}
        HACBR & 1 & 3.0$\times$10$^{-6}$ &  0.8687\\
             & 20 & 1.4$\times$10$^{-7}$& 0.9920 \\
        \cmidrule(r){1-4}
        HABR & 1 & 9.9$\times$10$^{-9}$ & 0.8443\\
             & 20 & 9.3$\times$10$^{-10}$& 0.9921\\
        \bottomrule
    \end{tabular}
    \caption{Total variance of the prior distribution and the ratio of between-group variance to total variance under different strategies 
and $ \beta $ on the CIFAR-10 dataset.}
    \label{tab: priorvar}
\end{table}

As shown in Table 2, within a certain range of $ \beta $, the ratio of between-group variance to 
total variance is negligible across all strategies. This indicates that when constraining the 
variance of the prior distribution, it is sufficient to limit the range of $\Sigma_{\phi(y)}$  while ignoring $\mu_{\phi(y)}$. 
For $\mu_{\phi(y)}$ and $\log \Sigma_{\phi(y)}$ of the distribution $q_{\phi}(z \mid y)$ corresponding to each label  $ y $  on the CIFAR-10 dataset, see \cref{Experiment Detail}.  

\section{Discussion}
\label{Discussion}
The application of block Flow requires prior knowledge associated with 
the dataset samples, typically in the form of labels $ y $. 
However, many datasets may not inherently include such information. 
For unlabeled datasets, clustering or similar methods can be utilized 
to annotate labels, thereby enabling the use of our approach.  

By leveraging labels $ y $, block Flow partitions the data distribution 
into blocks, which are then paired with blocks of the prior distribution. 
This approach effectively reduces intersections in forward trajectories, 
particularly cross-label intersections. 
It is anticipated that as the diversity of labels increases, 
the data blocks will become smaller and more numerous, 
further minimizing cross-label intersections. However, 
this also results in increased complexity in $ p(y) $. 

To implement block matching, we parameterize the prior distribution as a Gaussian mixture. However, it is inevitable that the ratio of between-group variance to total variance becomes extremely small, indicating that the means of the Gaussian components are very close to each other, resulting in significant overlap between the distributions. Consequently, the Gaussian mixture distribution closely resembles a single Gaussian distribution, failing to effectively model complex data distributions or capture the label-dependent characteristics of the data.  

In the hybrid information matching scheme, we attempted to capture the features of both samples $ x $ and labels $ y $, but the desired performance was not achieved. This may be attributed to the use of a small encoder to reduce computational cost, which limits the ability to capture complex features. Employing a larger encoder or leveraging pre-trained models could potentially address this issue.  

\section{Conclusion}
In this work, we propose block matching, 
a method to reduce intersections in forward trajectories by 
constructing a specialized coupling between the prior distribution 
and the data distribution using label $ y $. 
We establish the intrinsic relationship between the prior distribution 
and the curvature of the forward trajectories, providing a theoretical 
foundation for curvature control. We design multiple regularization 
strategies, all of which demonstrate competitive results with the same 
parameter scale. Our method has the potential to synergize with existing 
techniques, promising higher generation quality and faster sampling speeds.  

\section*{Impact Statement}
This paper presents work aimed at advancing the field of generative models, particularly in the context of flow-matching methods for image synthesis. Our proposed block matching technique and regularization strategies have the potential to improve the quality and efficiency of image generation, with applications in areas such as computer vision, medical imaging, and creative arts. While our research focuses on technical advancements, we acknowledge the broader societal implications of generative technologies, including ethical considerations related to data usage and the potential for misuse. We encourage responsible deployment and ongoing discussions to address these challenges.

\bibliographystyle{unsrt}  
\bibliography{references}  

\begin{thebibliography}{10}

\bibitem{sohl2015deep}
Jascha Sohl-Dickstein, Eric Weiss, Niru Maheswaranathan, and Surya Ganguli.
\newblock Deep unsupervised learning using nonequilibrium thermodynamics.
\newblock In {\em International conference on machine learning}, pages 2256--2265. PMLR, 2015.

\bibitem{ho2020denoising}
Jonathan Ho, Ajay Jain, and Pieter Abbeel.
\newblock Denoising diffusion probabilistic models.
\newblock {\em Advances in neural information processing systems}, 33:6840--6851, 2020.

\bibitem{song2020score}
Yang Song, Jascha Sohl-Dickstein, Diederik~P Kingma, Abhishek Kumar, Stefano Ermon, and Ben Poole.
\newblock Score-based generative modeling through stochastic differential equations.
\newblock {\em arXiv preprint arXiv:2011.13456}, 2020.

\bibitem{rombach2022high}
Robin Rombach, Andreas Blattmann, Dominik Lorenz, Patrick Esser, and Bj{\"o}rn Ommer.
\newblock High-resolution image synthesis with latent diffusion models.
\newblock In {\em Proceedings of the IEEE/CVF conference on computer vision and pattern recognition}, pages 10684--10695, 2022.

\bibitem{esser2024scaling}
Patrick Esser, Sumith Kulal, Andreas Blattmann, Rahim Entezari, Jonas M{\"u}ller, Harry Saini, Yam Levi, Dominik Lorenz, Axel Sauer, Frederic Boesel, et~al.
\newblock Scaling rectified flow transformers for high-resolution image synthesis.
\newblock In {\em Forty-first International Conference on Machine Learning}, 2024.

\bibitem{ho2022video}
Jonathan Ho, Tim Salimans, Alexey Gritsenko, William Chan, Mohammad Norouzi, and David~J Fleet.
\newblock Video diffusion models.
\newblock {\em Advances in Neural Information Processing Systems}, 35:8633--8646, 2022.

\bibitem{bar2024lumiere}
Omer Bar-Tal, Hila Chefer, Omer Tov, Charles Herrmann, Roni Paiss, Shiran Zada, Ariel Ephrat, Junhwa Hur, Guanghui Liu, Amit Raj, et~al.
\newblock Lumiere: A space-time diffusion model for video generation.
\newblock In {\em SIGGRAPH Asia 2024 Conference Papers}, pages 1--11, 2024.

\bibitem{kong2020diffwave}
Zhifeng Kong, Wei Ping, Jiaji Huang, Kexin Zhao, and Bryan Catanzaro.
\newblock Diffwave: A versatile diffusion model for audio synthesis.
\newblock {\em arXiv preprint arXiv:2009.09761}, 2020.

\bibitem{abramson2024accurate}
Josh Abramson, Jonas Adler, Jack Dunger, Richard Evans, Tim Green, Alexander Pritzel, Olaf Ronneberger, Lindsay Willmore, Andrew~J Ballard, Joshua Bambrick, et~al.
\newblock Accurate structure prediction of biomolecular interactions with alphafold 3.
\newblock {\em Nature}, pages 1--3, 2024.

\bibitem{vincent2011connection}
Pascal Vincent.
\newblock A connection between score matching and denoising autoencoders.
\newblock {\em Neural computation}, 23(7):1661--1674, 2011.

\bibitem{ramesh2022hierarchical}
Aditya Ramesh, Prafulla Dhariwal, Alex Nichol, Casey Chu, and Mark Chen.
\newblock Hierarchical text-conditional image generation with clip latents.
\newblock {\em arXiv preprint arXiv:2204.06125}, 1(2):3, 2022.

\bibitem{saharia2022photorealistic}
Chitwan Saharia, William Chan, Saurabh Saxena, Lala Li, Jay Whang, Emily~L Denton, Kamyar Ghasemipour, Raphael Gontijo~Lopes, Burcu Karagol~Ayan, Tim Salimans, et~al.
\newblock Photorealistic text-to-image diffusion models with deep language understanding.
\newblock {\em Advances in neural information processing systems}, 35:36479--36494, 2022.

\bibitem{chen2018neural}
Ricky~TQ Chen, Yulia Rubanova, Jesse Bettencourt, and David~K Duvenaud.
\newblock Neural ordinary differential equations.
\newblock {\em Advances in neural information processing systems}, 31, 2018.

\bibitem{song2021maximum}
Yang Song, Conor Durkan, Iain Murray, and Stefano Ermon.
\newblock Maximum likelihood training of score-based diffusion models.
\newblock {\em Advances in neural information processing systems}, 34:1415--1428, 2021.

\bibitem{lipman2023flowmatchinggenerativemodeling}
Yaron Lipman, Ricky T.~Q. Chen, Heli Ben-Hamu, Maximilian Nickel, and Matt Le.
\newblock Flow matching for generative modeling, 2023.

\bibitem{liu2022flowstraightfastlearning}
Xingchao Liu, Chengyue Gong, and Qiang Liu.
\newblock Flow straight and fast: Learning to generate and transfer data with rectified flow, 2022.

\bibitem{nguyen2023bellman}
Bao Nguyen, Binh Nguyen, and Viet~Anh Nguyen.
\newblock Bellman optimal step-size straightening of flow-matching models.
\newblock {\em arXiv preprint arXiv:2312.16414}, 2023.

\bibitem{lee2023minimizing}
Sangyun Lee, Beomsu Kim, and Jong~Chul Ye.
\newblock Minimizing trajectory curvature of ode-based generative models.
\newblock {\em arXiv preprint arXiv:2301.12003}, 2023.

\bibitem{murphy2012machine}
Kevin~P. Murphy.
\newblock {\em Machine Learning: A Probabilistic Perspective}.
\newblock MIT Press, 2012.

\bibitem{karras2020training}
Tero Karras, Miika Aittala, Janne Hellsten, Samuli Laine, Jaakko Lehtinen, and Timo Aila.
\newblock Training generative adversarial networks with limited data.
\newblock {\em Advances in neural information processing systems}, 33:12104--12114, 2020.

\bibitem{zhao2020differentiable}
Shengyu Zhao, Zhijian Liu, Ji~Lin, Jun-Yan Zhu, and Song Han.
\newblock Differentiable augmentation for data-efficient gan training.
\newblock {\em Advances in neural information processing systems}, 33:7559--7570, 2020.

\bibitem{jiang2021transgan}
Yifan Jiang, Shiyu Chang, and Zhangyang Wang.
\newblock Transgan: Two pure transformers can make one strong gan, and that can scale up.
\newblock {\em Advances in Neural Information Processing Systems}, 34:14745--14758, 2021.

\bibitem{cao2024spiking}
Jiahang Cao, Hanzhong Guo, Ziqing Wang, Deming Zhou, Hao Cheng, Qiang Zhang, and Renjing Xu.
\newblock Spiking diffusion models.
\newblock {\em IEEE Transactions on Artificial Intelligence}, 2024.

\bibitem{xu2023pfgm++}
Yilun Xu, Ziming Liu, Yonglong Tian, Shangyuan Tong, Max Tegmark, and Tommi Jaakkola.
\newblock Pfgm++: Unlocking the potential of physics-inspired generative models.
\newblock In {\em International Conference on Machine Learning}, pages 38566--38591. PMLR, 2023.

\bibitem{jia2024struct}
Nanshan Jia, Tingyu Zhu, Haoyu Liu, and Zeyu Zheng.
\newblock Structured diffusion models with mixture of gaussians as prior distribution, 2024.

\end{thebibliography}

\newpage
\appendix
\onecolumn
\section{Proof of Proposition}

\textbf{\cref{prop:v=0}.} If the probability distribution of $ x_c $ is a Dirac delta function, i.e.,  
$x_c \sim \delta(x - c),$ 
then:
\begin{equation}
    V((x_0, x_c))=0.
\end{equation}

\begin{proof}
At the final time $ t = 1 $, $  x_t = c $, and thus 
$  \mathbb{E}[c - x_0 \mid c] = \mathbb{E}[c - x_0] $, as $  c $  is a constant, the following holds:
\begin{align}
\mathbb{E}_{x_0 \sim p(x_0),x_c\equiv c} \left[ \left\| (x_c - x_0) - \mathbb{E}[x_c - x_0 \mid x_c]  \right\|^2_2 \right] 
&=\mathbb{E}_{x_0 \sim p(x_0)} \left[ \left\| (c-x_0) - \mathbb{E}[c-x_0 \mid c] \right\|^2_2 \right] \nonumber\\     
&= \mathbb{E}_{x_0 \sim p(x_0)} \left[ \left\| (c-x_0) - \mathbb{E}[c-x_0 ] \right\|^2_2 \right] \nonumber\\             
&=\text{Var}(x_0).
\end{align}
The integrand is bounded at $ t = 1 $, therefore:
\begin{align}
V((x_0, x_c)) &= \int_0^1 \mathbb{E}_{x_0 \sim p(x_0)} \left[ \left\| (c-x_0) - \mathbb{E}[c-x_0 \mid t c+(1-t) x_0] \right\|^2_2 \right] dt \nonumber\\ 
&= \int_{0}^{1-} \mathbb{E}_{x_0 \sim p(x_0)} \left[ \left\| (c-x_0) - \mathbb{E}[c-x_0 \mid t c+(1-t) x_0] \right\|^2_2 \right] dt \nonumber\\           
&= \int_{0}^{1-} \mathbb{E}_{x_0 \sim p(x_0)} \left[ \left\| (c-x_0) - \mathbb{E}[c-x_0 \mid x_0] \right\|^2_2 \right] dt \nonumber\\            
&= \int_{0}^{1-} \mathbb{E}_{x_0 \sim p(x_0)} \left[ \left\| (c-x_0) - (c-x_0)  \right\|^2_2 \right] dt \nonumber\\            
&=0.
\end{align}
\end{proof}

\textbf{\cref{prop:v<var}.} For any joint distribution $p(x_0, x_1)$, the following holds:  
\begin{equation}
V((x_0, x_1))\leq \left( \sqrt{\text{Var}(x_1)} + \sqrt{\text{Var}(x_0)} \right)^2.    
\end{equation}

In particular, when $ x_0 $ and $ x_1 $ are independent, we have:  
\begin{equation}
V((x_0, x_1))\leq \text{Var}(x_1) + \text{Var}(x_0).    
\end{equation}

\begin{proof}
Given:
\begin{equation}
V((x_0, x_1)) = \int_0^1 \mathbb{E}_{(x_0,x_1)\sim p(x_0, x_1)} \left[ \left\| (x_1 - x_0) - \mathbb{E}[x_1 - x_0 \mid x_t] \right\|^2_2 \right] dt,
\end{equation}

where $ x_t = t x_1 + (1 - t) x_0 $.

\begin{align}
    V((x_0, x_1))
            &= \int_0^1 \mathbb{E}_{x_t,(x_0,x_1)\sim p(x_0, x_1)} \left[ \left( (x_1 - x_0) - \mathbb{E}[x_1 - x_0 \mid x_t] \right)^2_2 \mid x_t\right] dt 
\end{align}

By the definition of conditional variance:
\begin{equation}
\mathbb{E}_{x_t,(x_0,x_1)\sim p(x_0, x_1)} \left[ \left( (x_1 - x_0) - 
\mathbb{E}[x_1 - x_0 \mid x_t] \right)^2_2 \mid x_t\right] 
= \mathbb{E}_{x_t} \left[ \text{Var}(x_1 - x_0 \mid x_t) \right].    
\end{equation}

Thus:
\begin{equation}
V((x_0, x_1)) = \int_0^1 \mathbb{E}_{x_t} \left[ \text{Var}(x_1 - x_0 \mid x_t) \right] dt.
\end{equation}

By the law of total variance:
\begin{equation}
\text{Var}(x_1 - x_0) = \mathbb{E}_{x_t} \left[ \text{Var}(x_1 - x_0 \mid x_t) \right] + \text{Var}_{x_t} \left( \mathbb{E}[x_1 - x_0 \mid x_t] \right).    
\end{equation}

Therefore:
\begin{align}
    V((x_0, x_1)) &= \int_0^1  \text{Var}(x_1 - x_0) - \text{Var}_{x_t} \left( \mathbb{E}[x_1 - x_0 \mid x_t] \right)dt  \nonumber\\
            &\leq  \int_0^1  \text{Var}(x_1 - x_0)dt \nonumber\\
            &=\text{Var}(x_1 - x_0)\nonumber\\
            &=  \text{Var}(x_1) + \text{Var}(x_0) - 2\text{Cov}(x_1, x_0)\nonumber\\
            &\leq\text{Var}(x_1) + \text{Var}(x_0) + 2 \sqrt{\text{Var}(x_1)\text{Var}(x_0)}\nonumber\\
            &=\left( \sqrt{\text{Var}(x_1)} + \sqrt{\text{Var}(x_0)} \right)^2.
\end{align}
In particular, when $ x_0 $ and $ x_1 $ are independent, we have:  
\begin{equation}
\text{Cov}(x_1, x_0)=0.    
\end{equation}

Therefore:
\begin{align}
    V((x_0, x_1))&\leq  \text{Var}(x_1) + \text{Var}(x_0) - 2\text{Cov}(x_1, x_0)\nonumber\\
            &=\text{Var}(x_1) + \text{Var}(x_0).
\end{align}
\end{proof}

\section{Experiment Detail}
\label{Experiment Detail}
In the HACBR strategy, the variance of $q_{\phi}(z \mid x, y)$ is set to 1 to prevent the collapse of the prior distribution. 
In other words, the regularization term $R(\phi)=D_{\text{KL}} \left( q_{\phi}(z \mid x, y) || p_0(z) \right)$ is essentially KL divergence between $ \mathcal{N}\left(\mu_{\phi(x,y)},\bm{I}\right)  $ and $ \mathcal{N}\left(\mu_{\phi(y)} , \Sigma_{\phi(y)}\right) $.

\cref{mu var of FABR} and \cref{mu var of FANR} present $\mu_{\phi(y)}$ and $\log \Sigma_{\phi(y)}$ of $q_{\phi}(z \mid y)$ for each label $ y $ in block flow(FABR) 
on the CIFAR-10 dataset across different $ \beta $.
\begin{table}[h]
\begin{center}
\begin{tabular}{lllllll}
\toprule
\multirow{2}{*}{FABR} & \multicolumn{2}{l}{$\beta=0.1$}     & \multicolumn{2}{l}{$\beta=1$}       & \multicolumn{2}{l}{$\beta=20$}       \\ \cline{2-7} 
                      & $\mu_{\phi(y)}$                  & $\log \Sigma_{\phi(y)}$ & $\mu_{\phi(y)}$                  & $\log \Sigma_{\phi(y)}$ & $\mu_{\phi(y)}$                   & $\log \Sigma_{\phi(y)}$ \\ \midrule
1                     & 5.57$\times 10 ^{-5}$  & -1.0654    & 2.99$\times 10 ^{-4}$  & -0.1357    & 1.38$\times 10 ^{-5}$   & -0.0068    \\
2                     & -3.10$\times 10 ^{-4}$ & -1.3465    & -2.13$\times 10 ^{-4}$ & -0.1786    & -1.27$\times 10 ^{-5}$  & -0.0089    \\
3                     & 8.53$\times 10 ^{-5}$  & -1.2003    & 1.70$\times 10 ^{-4}$  & -0.1489    & 9.33$\times 10 ^{-6}$   & -0.0074    \\
4                     & -2.54$\times 10 ^{-4}$ & -1.2769    & 6.70$\times 10 ^{-5}$  & -0.1582    & -3.38$\times 10 ^{-6}$  & -0.0084    \\
5                     & 1.49$\times 10 ^{-4}$  & -1.2892    & 3.06$\times 10 ^{-5}$  & -0.1565    & 1.05$\times 10 ^{-6}$ & -0.0078    \\
6                     & -2.24$\times 10 ^{-4}$ & -1.2874    & -7.17$\times 10 ^{-5}$ & -0.1698    & -3.80$\times 10 ^{-6}$  & -0.0084    \\
7                     & -5.10$\times 10 ^{-6}$ & -1.4465    & -1.77$\times 10 ^{-4}$ & -0.1771    & -8.74$\times 10 ^{-6}$  & -0.0088    \\
8                     & -1.81$\times 10 ^{-4}$ & -1.3291    & -1.35$\times 10 ^{-4}$ & -0.1718    & -7.80$\times 10 ^{-6}$  & -0.0086    \\
9                     & 1.92$\times 10 ^{-5}$  & -1.1292    & 2.08$\times 10 ^{-4}$  & -0.1433    & 9.60$\times 10 ^{-6}$ & -0.0071    \\
10                    & -4.03$\times 10 ^{-4}$ & -1.4018    & -2.60$\times 10 ^{-4}$ & -0.1864    & -1.48$\times 10 ^{-5}$  & -0.0093    \\ \bottomrule
\end{tabular}
\end{center}
\caption{$\mu_{\phi(y)}$ and $\log \Sigma_{\phi(y)}$ of FABR 
on the CIFAR-10 dataset.}
\label{mu var of FABR}
\end{table}

\begin{table}[h]
\begin{center}
\begin{tabular}{lllll}
\toprule
\multirow{2}{*}{FANR} & \multicolumn{2}{l}{$\beta=0.1$}                         & \multicolumn{2}{l}{$\beta=1$} \\ \cline{2-5} 
                      & $\mu_{\phi(y)}$     & $\log \Sigma_{\phi(y)}$ & $\mu_{\phi(y)}$  & $\log \Sigma_{\phi(y)}$    \\ \midrule
1                     & -0.0013   & -0.1918                                     & -0.0087    & -4.37$\times 10 ^{-6}$      \\
2                     & -0.0011   & -0.1997                                     & -0.0087    & -4.37$\times 10 ^{-6}$        \\
3                     & -0.0015   & -0.1851                                     & -0.0087    & -4.37$\times 10 ^{-6}$        \\
4                     & -0.0014   & -0.189                                      & -0.0087    & -4.37$\times 10 ^{-6}$        \\
5                     & -0.0014   & -0.1891                                     & -0.0087    & -4.37$\times 10 ^{-6}$        \\
6                     & -0.0014   & -0.1904                                     & -0.0087    & -4.37$\times 10 ^{-6}$        \\
7                     & -0.0017   & -0.1776                                     & -0.0087    & -4.37$\times 10 ^{-6}$        \\
8                     & -0.0013 & -0.1935                                     & -0.0087    & -4.37$\times 10 ^{-6}$        \\
9                     & -0.0014   & -0.1897                                     & -0.0087    & -4.37$\times 10 ^{-6}$        \\
10                    & -0.0011   & -0.1990                                      & -0.0087    & -4.37$\times 10 ^{-6}$        \\ \bottomrule
\end{tabular}
\end{center}
\caption{$\mu_{\phi(y)}$ and $\log \Sigma_{\phi(y)}$ of FANR 
on the CIFAR-10 dataset.}
\label{mu var of FANR}
\end{table}

\begin{figure}[h]
\begin{center}
\centerline{\includegraphics[width=0.8\columnwidth]{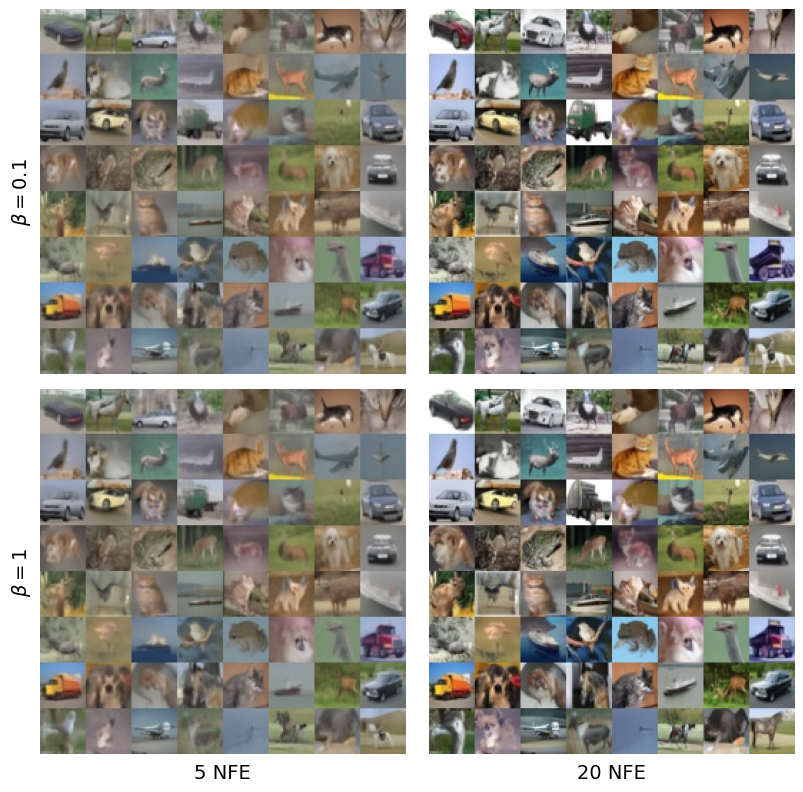}}
\caption{Results of block flow (FANR) with the $L_2 $ norm on CIFAR-10 datase}
\label{cifar10_norm2}
\end{center}
\end{figure}

\begin{figure}[h]
\begin{center}
\centerline{\includegraphics[width=0.8\columnwidth]{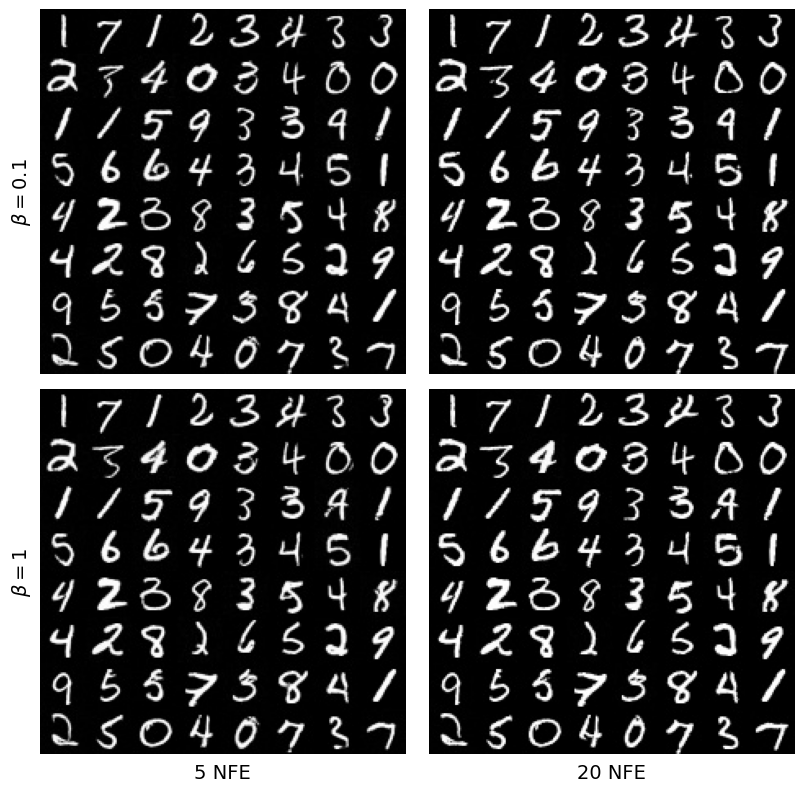}}
\caption{Results of block flow (FANR) with the $L_\infty $ norm on MNIST datase}
\label{mnist_norminf}
\end{center}
\end{figure}

FANR defaults to using the $L_1$ norm.However, the $L_\infty $ norm and $L_2$ norm are also viable alternatives.
The sampling results using the Euler method are shown in Fig.~\ref{cifar10_norm2} and Fig.~\ref{mnist_norminf}.

\begin{figure}[h]
\begin{center}
\centerline{\includegraphics[width=0.8\columnwidth]{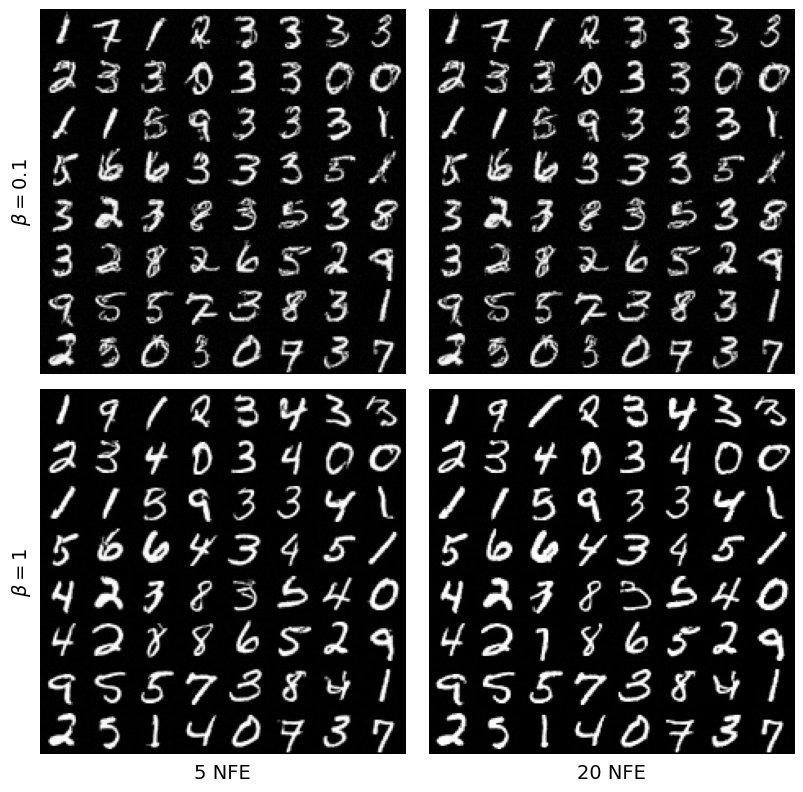}}
\caption{Results of block flow (HABR) without label on MNIST datase}
\label{mnist_HABR_without label}
\end{center}
\end{figure}

\begin{figure}[h]
\begin{center}
\centerline{\includegraphics[width=0.8\columnwidth]{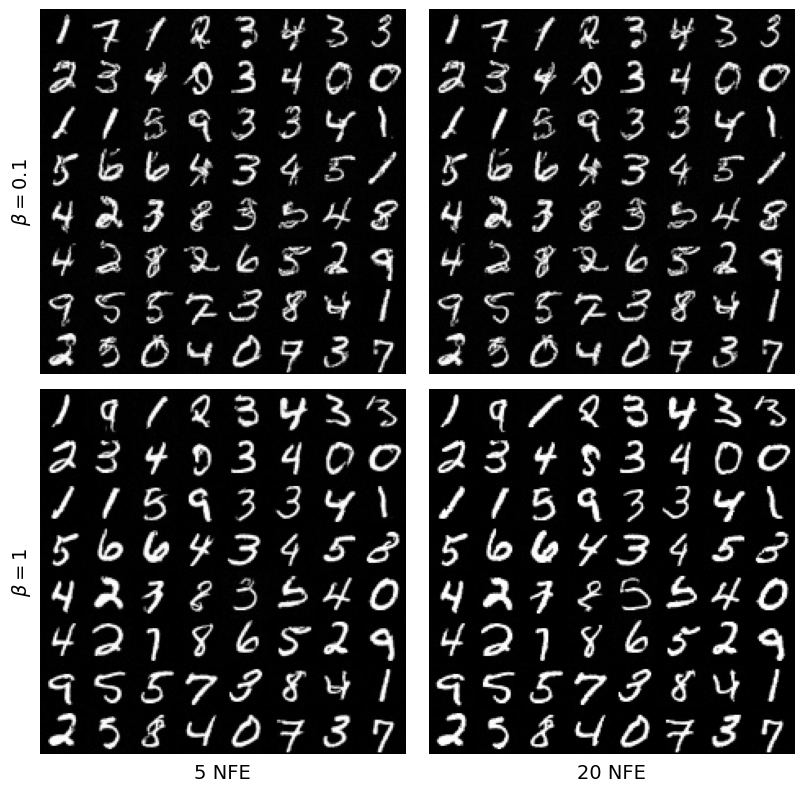}}
\caption{Results of block flow (HACBR) without label on MNIST datase}
\label{mnist_HACBR_without label}
\end{center}
\end{figure}

In the experiments, the decoder defaults to using label $ y $ for conditional generation, 
consistent with the encoder. However, it is also feasible for the decoder to operate without 
label $ y $. The results are shown in Fig.~\ref{mnist_HABR_without label} and Fig.~\ref{mnist_HACBR_without label}.

\end{document}